\pdfoutput=1

\documentclass[11pt]{article}
\usepackage[final]{gwc}

\usepackage{times}
\usepackage{latexsym}
\usepackage[T1]{fontenc}
\usepackage[utf8]{inputenc}
\usepackage{inconsolata}
\usepackage{graphicx}
\usepackage{booktabs}
\usepackage{multirow}
\usepackage{caption}
\usepackage{subcaption}
\usepackage{makecell}
\usepackage{lscape}
\usepackage{float}
\usepackage{amsmath}

\newcommand{\textdef}[1]{\textit{#1}}
\newcommand{\word}[1]{\textit{``#1''}}
\newcommand{\W}[0]{\textnormal{[\textsc{target}]}}
\newcommand{\V}[0]{\textnormal{[\textsc{relatum}]}}

\title{Misalignment of Semantic Relation Knowledge \\ between WordNet and Human Intuition}

\author{
    Zhihan Cao \\
    Institute of Science Tokyo \\
    \texttt{cao.z.ab@m.titech.ac.jp}
    \And
    Hiroaki Yamada \\
    Institute of Science Tokyo \\
    \texttt{yamada@comp.isct.ac.jp}
    \AND
    Simone Teufel \\
    University of Cambridge \\
    \texttt{simone.teufel@cam.ac.uk}
    \And
    Takenobu Tokunaga \\
    Institute of Science Tokyo \\
    \texttt{take@c.titech.ac.jp} \\
}

\begin{document}

\maketitle

\begin{abstract}
    WordNet provides a carefully constructed repository of semantic relations, created by specialists. 
    But there is another source of information on semantic relations, the intuition of language users. 
    We present the first systematic study of the degree to which these two sources are aligned.
    Investigating the cases of misalignment could make proper use of WordNet and facilitate its improvement.
    Our analysis which uses templates to elicit responses from human participants, reveals a general misalignment of semantic relation knowledge between WordNet and human intuition.
    Further analyses find a systematic pattern of mismatch among synonymy and taxonomic relations~(hypernymy and hyponymy), together with the fact that WordNet path length does not serve as a reliable indicator of human intuition regarding hypernymy or hyponymy relations.
\end{abstract}


\section{Introduction}
Semantic relations represent how the senses of two lexical items are related.
These relations structure the vocabulary of natural languages~\citep{Miller_1991a,semantic_priming,semantics}, making them essential for both human language comprehension and production.
On the practical side, the performance of a wide range of natural language processing (NLP) tasks improves when incorporating information about semantic relations, including text simplification, paraphrasing, natural language inference, and discourse analysis~\citep{Tatu_2005, Madnani_2010, Glavas_2015, Alamillo_2023}.
Therefore, semantic relations are not only important in analyzing languages but also a crucial piece of lexico-semantic information for NLP even in the current era of large language models.

In order to afford the analysis of the lexical semantics and the evaluation of whether large language models properly recognize semantic relations, we need a good resource of semantic relations at hand.
Such a resource should contain only valid items and have as wide a coverage as possible.

Experts' introspection can guarantee the validity.
An example of such resources is WordNet~\citep{Miller_1995}.
The core object of WordNet is the \textdef{synset}, a set of synonymous lexical items which represents a word sense.
Semantic relations are then defined on a pair of synsets\footnote{Antonymy is the exception. It is defined on a pair of synset-disambiguated lexical items, or \textdef{lemmas} in the WordNet terminology.}.
WordNet was constructed in the 1980s by lexicographers and successively updated until 2006, with version 3.1 as the final release.

WordNet encompasses diverse semantic relations: hypernymy, hyponymy, holonymy, meronymy, antonymy, synonymy, and some others.
However, a well-known drawback of WordNet (other than the fact that the project has stopped and has therefore not been updated for years) is that semantic relations are not treated equally;
synsets are constructed on the basis of synonymy and 
the WordNet hierarchical structure is built by hypernymy and hyponymy links between the synsets.
Other relations, such as antonymy, holonymy, and meronymy, are documented less.
Hence, there is an imbalance in the number of synset pairs in different semantic relations~\footnote{Among all 82,115 nominal synsets, more than 90\% are linked to at least one hypernym.
However, only 25\% of synsets have at least one holonym.
For meronymy, it is merely 12\%. }, indicating the coverage of WordNet is limited.

Language users' intuition is an orthogonal resource of semantic relation knowledge that we can tap into.
A body of research has shown that such intuition can be used to augment or modify WordNet. 
For example, \citet{Veale_2008} introduce the modifier--modifiee relation into WordNet, which is a relation between an adjective and a noun and expresses their cultural association.
To do so, they mine real-world similes online, based on the construction \word{as \textnormal{\textsc{adjective}} as \textnormal{\textsc{noun}}}.
Word sense disambiguation is performed afterwards in order to establish connections to WordNet.
Their evaluation compares the modifier adjectives mined online with adjectives extracted from the WordNet glosses.
They find improved performance in determining the sentiment of the modifiee noun when using the adjectives mined.
\citet{wordnet2019, wordnet2020} also use human intuition to improve WordNet. 
Their methodology is based on collecting explicit feedback from WordNet users.
Using that methodology, they have detected missing or wrong lexical items in a synset, lack of synsets, lack of relations, and inappropriate relations.

Previous approaches offer refined but pointwise modifications:  data-mining methods are prone to frequency biases, often resulting in suboptimal performance for low-frequency terms;  feedback-based approaches rely on incidental discoveries by a self-selected group of NLP practitioners. 
In order to build the best possible repository of semantic relations, it becomes efficient and effective if we integrate refined and systematic approaches.

Developing systematic augmentations requires a holistic understanding of WordNet.
As a preliminary step, this study aims to achieve such an understanding by carefully investigating the alignment between language users' intuition and expert opinions regarding semantic relations. 
Our results reveal a general misalignment across various semantic relations, with distinct patterns emerging from deeper analyses. 
The data collected in this work is made available at \url{https://github.com/hancules/HumanElicitedTriplets}.

\section{Method}
The fundamental question of the present study is to what extent semantic relation knowledge documented in WordNet
\footnote{We use the modified WordNet version by \citeauthor{wordnet2020}}
aligns with the knowledge held by language users.
We include six relations: hypernymy~(HYP), hyponymy~(HPO), holonymy~(HOL), meronymy~(MER), antonymy~(ANT), and synonymy~(SYN).
We study the alignment separately for each relation. 

\subsection{Elicitation}
The main unit we work with is the triplet consisting of a target word $w$, a relation $r$ and a relatum $v$. 
The sentence \word{an orange is kind of a fruit} that expresses a hypernymy relation would then be translated into (\word{orange}, HYP, \word{fruit}).
We use the triplets to compare the information in WordNet and the relational knowledge of language users. 
We collect triplets from language users by elicitation, a well-established methodology in linguistics~\citep{elicitation}~(the procedure will be explained in Section~\ref{sec:data}).

\subsection{Match Status}
%
%
Some elicited triplets already exist in WordNet with the same direct relation~(i.e. not related through transitivity); such elicited triplets are called \textbf{matched triplets}.
Another type of elicited triplets is the \textbf{missing triplet}, where no direct relation between the target word and relatum is documented, although both individually exist in WordNet. 
It can also happen that the target word and relatum in elicited triplets are both found in WordNet but in a different relation.
We name these \textbf{mismatched triplets}.
The existence of matched triplets confirms our confidence in WordNet's information, and the missing and mismatched triplets are potential resources for the improvement of WordNet.


\subsection{Analysis Objectives}
We start with two analyses that intend to capture the general picture of alignment.
If WordNet aligns with language users' intuition, there should be more matched triplets and fewer missing and mismatched triplets.
Our first analysis investigates the distribution of three match statuses.

Some triplets may be elicited commonly from language users.
We can calculate the \textdef{elicitation frequency} for each triplet.
Elicitation frequency is an indicator of how intuitive the corresponding triplet is to the language users. 
A good alignment should also result in a monotonic increase relationship between the elicitation frequency and the intuitiveness of matched triplets.
Our second analysis looks at whether highly intuitive triplets are more likely to be documented in WordNet.

For mismatched triples, we measure \textdef{mismatch likelihood}, which indicates how likely an elicited triplet is documented as a different relation in WordNet.
By comparing mismatch likelihoods of different relations for each elicited relation, we can figure out whether it occurs particularly for certain relations.
One possible reason for a high mismatch likelihood is that words could be polysemous, where the senses are related to each other.
The related nature between senses introduces a high mismatch likelihood between certain elicited and documented relations.
For example, if a word is a metonym, as it can form both synonymy and meronymy with another word, a high mismatch likelihood between synonymy and meronymy might be observed.

For transitive relations such as hypernymy and hyponymy, we extend the definition of matched triplets by considering indirect relations through transitivity in WordNet. 
By including those \textdef{indirectly matched triplets}, we study the path length between the target word and relatum.
The reason why we are interested in the path length is that the path length is considered the proxy of their semantic similarity~\citep{Resnik_1995}.
Various metrics explicitly or implicitly incorporate the path length in WordNet to capture the semantic similarity~\citep{Wu_1994, Resnik_1995, Fellbaum_1998}.
However, questions remain about whether the path length aligns with human intuition.
We calculate the correlation between the path length and the elicitation frequency.
Through this analysis, we can gain insights into the relation between the WordNet structure and human intuition.

\section{Data}
\label{sec:data}
\begin{table*}[htpb]
    \centering
    \begin{tabular}{c|rcrl}
    \toprule
    Relation &  \# Target Words &  \# Templates & \# Triplets & (\# Hapaxes) \\ 
    \midrule
    HYP  
    & 713  & 7 & 11,739 & (6,329) \\ 
    HPO  & 319  & 4 & 5,721 & (3,646) \\ 
    HOL  & 195  & 7 & 3,496 & (1,870) \\ 
    MER  & 146  & 6 & 2,997 & (1,568) \\ 
    ANT  & 105  & 9 & 1,447 & (804) \\ 
    SYN  & 218  & 7 & 3,094 & (1,589) \\ 
    \midrule
    TOTAL &\makecell[r]{1,696\\(1,304 unique)}  & 40  & 28,494 & 15,806 \\ 
    \bottomrule
    \end{tabular}
    \caption{Target words, templates, and the numbers of elicited triplets, with the numbers of hapaxes in brackets.}
    \label{tab:stats}
\end{table*}

\subsection{Template-based Elicitation}
The elicitation is carried out using templates~\citep{Ettinger_2020}.
We first verbalize a relation by a template, such as a hypernymy template \word{a {\W} is a type of {\V}}.
After {\W} is specified, the elicitation can be carried out as a cloze task: 
participants are asked to fill in slot {\V} with up to five words as relata, and we construct elicited triplets.

For all relations, we hand-craft templates.
Particularly for hypernymy and hyponymy, we design some of our templates based on lexico-syntactic patterns by \citet{Roller_2018}~\footnote{Their patterns are extracted in the same fashion as~\citet{Hearst_1992}.}.
Such templets for hypernymy are
\word{a {\W} is a kind of a {\V}} and
\word{a {\W} is a specific case of a {\V}} while for hyponymy, \word{a {\W}, such as a {\V}}.
Examples of other templates follow.
\begin{itemize}
    \item[] \textbf{HYP:} the word {\W} has a more specific sense than the word {\V},
    \item[] \textbf{HPO:} the word {\W} has a more general sense than the word {\V},
    \item[] \textbf{HOL:} a {\W} is contained in a {\V},
    \item[] \textbf{MER:} a {\W} contains a {\V},
    \item[] \textbf{ANT:} a {\W} is the opposite of a {\V},
    \item[] \textbf{SYN:} a {\W} is similar to a {\V}.
\end{itemize}

In total, we use nine templates for antonymy, seven for hypernymy, holonymy, and synonymy, six for meronymy, and four for hyponymy.
The full list of templates can be found in Appendix~\ref{app:templates}.

\subsection{Target Words}
\label{sec:target word}
We need target words to elicit relata from participants.
Proper target words need at least one possible relatum for the relation of interest.
Therefore, target words cannot be randomly sampled.

We exploit two existing human-confirmed corpora of triplets~\citep{hyperlex, Exbattig} to obtain valid target words.
We use only target words from these two resources but do not use their triplets as is.
This is because
1) none of them include all five relations of interest and
2) their size is too limited~(1,347 triplets in total).

We first remove triplets from the above corpora where either target word or relatum is not a noun or is not in the intersection of vocabularies of  BERT~\citep{bert}, RoBERTa~\citep{roberta}, and ALBERT~\citep{albert}\footnote{This is a design decision made in conjunction with an experiment reported in \citet{Cao_2024}.}.
Target words are then extracted from the resulting triplets, and augmented as follows.
For symmetric relations antonymy and synonymy, we extract both target words and relata in the current triplets as the target words for elicitation.
For each of the other relations, we extract the target words in triplets of the relation and the relatum in the triplets of the reverse relation.
Meronymy and holonomy are reversed to each other, and so are hyponymy and hypernymy. 
We remove any duplicates from the extracted target words.

This procedure results in 713 target words for hypernymy, 319 for hyponymy, 195 for holonymy, 146 for meronymy, 105 for antonymy, and 218 for synonymy\footnote{This amounts to 1,304 unique target words; note that some target words are associated with more than one relation.}.
These target words and templates yield a total of 10,979 task sentences used in the elicitation experiment.

\subsection{Collection of Elicited Triplets}
We use the Amazon Mechanical Turk~(MTurk) crowdsourcing platform in order to collect relata from language users. 
Participants are restricted to those who have the Mturk Master qualification and whose answers are approved more than 500 times at an approval rate beyond 95\%, and who additionally currently live in either the United States, the United Kingdom, Australia, or Canada.

We split the 10,979 task sentences into 276 subsets of around 37.8 sentences on average, making sure that no subset contains more than one sentence with the same relation and the same target word.
We collected responses from four participants for each subset.

In total, 48 qualified participants were recruited.
We explicitly instructed participants that they could use nouns and could not use multi-word expressions.
The time limit for responding to each sentence is three minutes.
Participants each answered 22 subsets on average. 
We collected 30,193 elicited triplets.

\subsection{Identification of Match Status}
The identification of the documented relation for an elicited triplet follows this procedure.
For all synsets of the target word and relatum, we first check if they are documented directly in any relation of interest.
If none, the triplet is identified as a missing triplet.
If the documented relation matches the elicited relation, it is then identified as a matched triplet.
In all other cases, the triplet is identified as a mismatched triplet.

Due to our treatment of word senses, some elicited triplets may include word pairs that stand in multiple relations in WordNet.
However, it is very rare and there are only 456 out of 30,193 elicited triplets.
We exclude such triplets.
We also exclude triplets whose relatum can not be found as a noun in WordNet~(there are 1,243 of these).
This results in 28,494 triplets, out of which 12,688 are hapaxes, observed only once (cf. Table~\ref{tab:stats}).

Although each template is designed specifically for only one relation, in reality, it is possible that the template might unintentionally evoke another relation for participants, according to their interpretation.
In such case, the relata elicited may not be related to the target word in the intended relation, resulting in invalid relata;
for each relation, we therefore measure the association between the distributions of relata coming from different templates.
The higher the association is, the more confidently we can say that the templates successfully express that relation.

To calculate the association among templates within each relation, we use two association metrics: \citeauthor{Cramer_1946}'s $V$~(\citeyear{Cramer_1946}) and the generalized Jensen-Shannon divergence (GJSD)~\citep{gjsd}.
\citeauthor{Cramer_1946}'s $V$ quantifies the association between multiple nominal samples.
It ranges from 0 to 1, where higher means more strongly associated. 
GJSD extends the Jensen-Shannon divergence so that it is able to compare multiple distributions.
GJSD ranges from 0 (similar) to 1 (dissimilar).

\begin{table}[htbp]
    \centering
    \begin{tabular}{l|cc}
    \toprule
    Relation & GJSD & \citeauthor{Cramer_1946}'s $V$ \\
    \midrule
    HYP	& 0.22 & 0.49 \\
    HPO	& 0.24 & 0.66 \\
    HOL	& 0.22 & 0.49 \\
    MER	& 0.21 & 0.52 \\
    ANT	& 0.21 & 0.43 \\
    SYN	& 0.22 & 0.47 \\
    \bottomrule
    \end{tabular}
    \caption{Association between templates per relation.}
    \label{tab:association}
\end{table}
Table~\ref{tab:association} shows the average metric scores over different target words per relation.
We observe that the average GJSD are around 0.20 across relations, indicating a high similarity of relata distributions that come from different templates within a relation.
The mean \citeauthor{Cramer_1946}'s $V$ are above 0.40 for all relations, which is generally interpreted as a strong association~\cite{cramerv, Akoglu_2018}.
We conclude that the templates within relations overall express the same relation.

\subsection{Metrics}
\subsubsection{Elicitation Frequency}
The elicitation frequency $\mathcal{F}$ of a triplet $(w,r,v)$ is defined as follows.
\begin{equation}
\label{eq:hed}
    \mathcal{F}\left(w, r, v\right) = 
    \frac{f\left(w,r,v\right)}
    {\sum_i{f\left(w,r,v_i\right)}}
\end{equation}
where $f(w,r,v)$ is the number of times $v$ was elicitated in relation $r$ to $w$ across templates. 

One of our analysis objectives is to observe whether highly intuitive triplets are more likely documented.
To do so, we create a curve of the elicitation frequencies and match rates.
Match rates are obtained as follows.
For a relation $r$ and a frequency threshold in a range of $[0,1]$, we retain triplets $(w,r,v)$ that have an elicitation frequency above it, and then the match rate is the proportion of matched triplets among them.
The curve of elicitation frequency and match rate is then generated by plotting each threshold and the match rate.

\subsubsection{Mismatch Likelihood}
The mismatch likelihood $\mathcal{L}$ measures to what extent an elicited relation $r$ is likely to be documented as a different relation $s$ in WordNet.
It is the normalized sum of frequencies of mismatched triplets whose elicited relation is $r$ and documented relation is $s$.
$T^r_s$ denotes a set of such mismatched triplets.
A mismatch likelihood of $s$ given $r$ is defined as follows.
\begin{align}
\label{eq:misma}
    g\left(s; r\right) = \sum_{\left(w, r, v\right) \in T^r_s}{
    \mathcal{F}\left(w, r, v\right)} \\
    \mathcal{L}\left(s; r\right) = \frac{ g\left(s; r\right)}{
    \sum_{R\setminus {\{t\}}}{g\left(t; r\right)}
    }
\end{align}
where $R\setminus {\{t\}}$ denotes the relation set with the relation $t$ excluded.
The mismatch likelihood is defined within the interval $[0,1]$.
For a given elicited relation, the mismatch likelihoods of the remaining five relations sum to one.
It enables a comparison of the five relations, allowing us to find out for which documented relation, the mismatch is most likely to occur.

\section{Results}
\subsection{Distribution of Triplet Categories}
\begin{figure*}[ht!]
    \centering
    \includegraphics[width=0.8\linewidth]{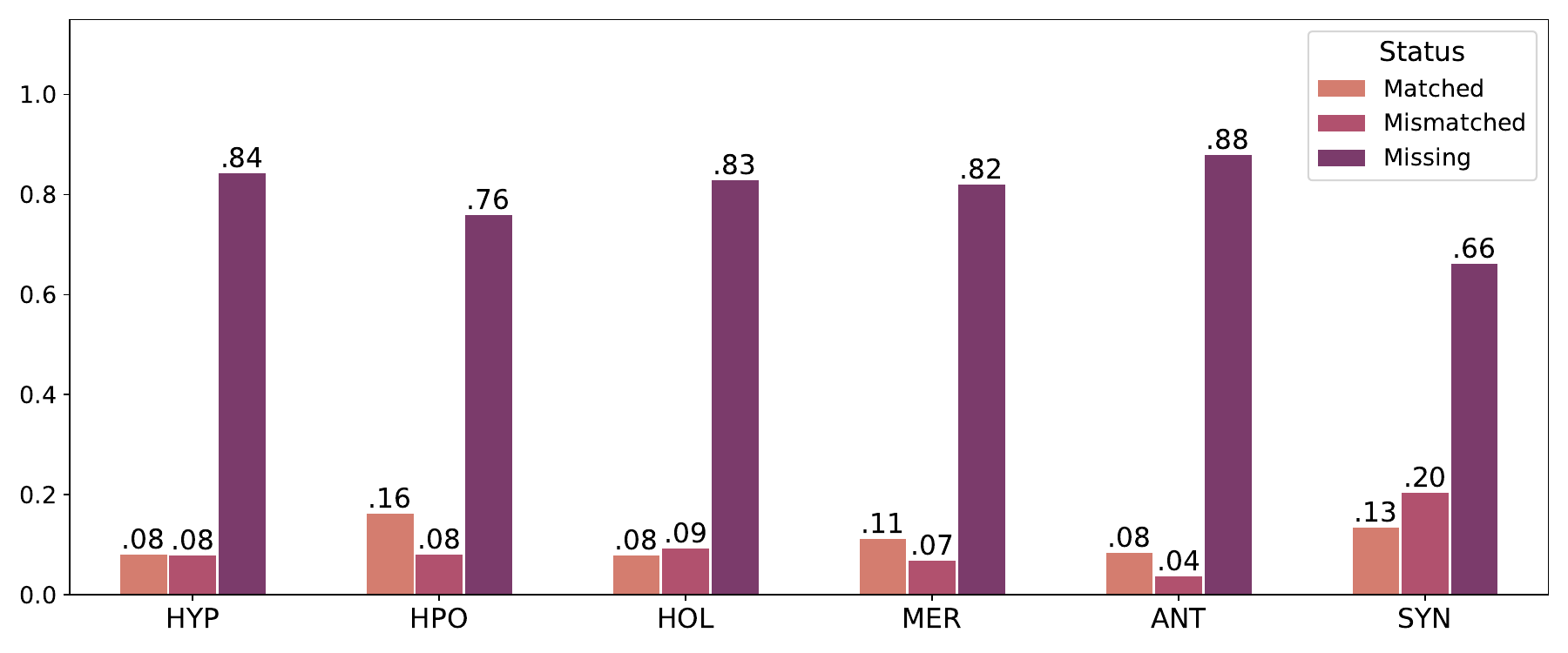}
    \caption{Match status distribution per relation.}
    \label{fig:catdist}
\end{figure*}
Figure~\ref{fig:catdist} displays the distributions of match statuses for each relation. 
Missing triplets dominate: more than 60\% of triplets are not found in WordNet.
This tendency holds for both hapax and non-hapax triplets\footnote{The distribution of hapax and non-hapax triplets can be found in Appendix~\ref{app:distinguished-dist}.}.
It indicates a large misalignment between language users and WordNet concerning semantic relation knowledge.

There are more matched than mismatched triplets for hyponymy and meronymy, while they are comparable for hypernymy and holonymy.
In addition, the proportion of matched triplets is particularly low for hypernymy and holonymy, both below 10\%.
Hypernymy and holonymy are two relations that require abstraction for the generic form (hypernym) and for the whole (holonymy).
We might hypothesize that abstraction, a cognitively expensive process, pushes the proportion of matched triples lower.

Antonymy has the lowest mismatch rate of 4\%.
It shows a different behavior from synonymy, which has a very high mismatch rate at 20\%.
Antonymy triplets are most unlikely, whereas synonymy triplets are most likely, to be mismatched as other relational triplets.
This result can be explained by their definition.
Antonymy is the only relation discussed here that is based on mutual exclusion between the target word and relatum.
On the other hand, synonymy is based on inclusion, which is a property that hypernymy, hyponymy, holonymy, and meronymy also share to different degrees~\citep{Joosten_2010}.
Hence, there is a clear line between antonymy and all other five relations, resulting in a low mismatch rate of antonymy.

\subsection{Dynamics of Elicitation Frequency and Match Rate}
\begin{figure*}[htbp]
    \centering
    \includegraphics[width=\linewidth]{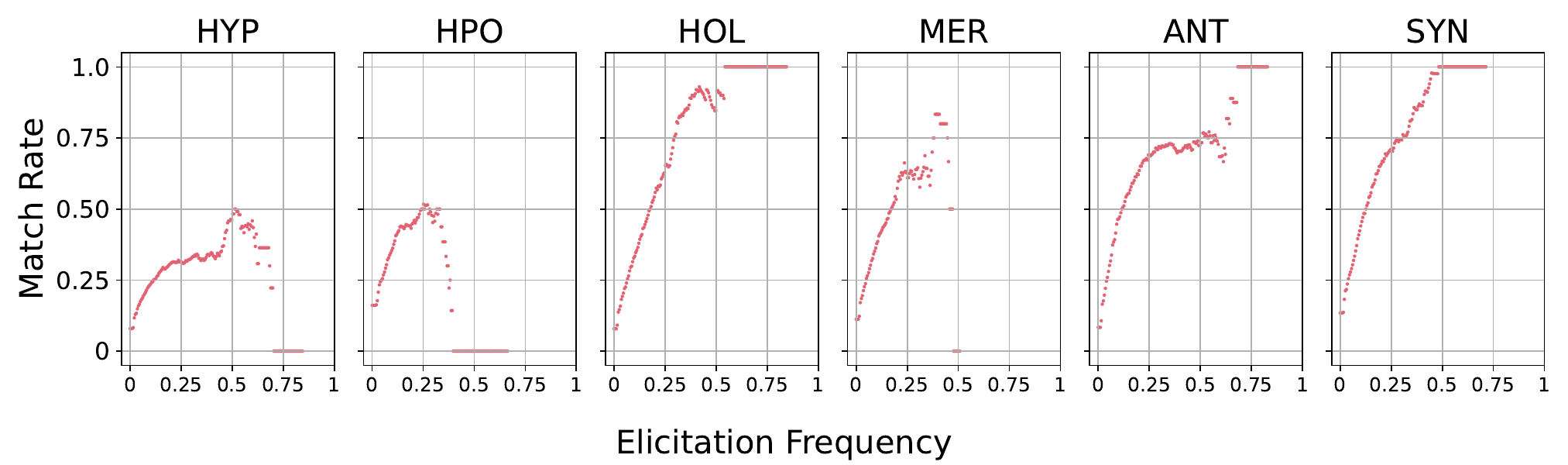}
    \caption{Human elicitation frequency vs. match rate.}
    \label{fig:matchrate}
\end{figure*}
Let us consider how match rates are related to the elicitation frequency (intuitiveness) of triples.
Figure~\ref{fig:matchrate} shows the relation between the elicitation frequency and the match rate per relation.
If the more intuitive triplets are more likely documented in WordNet, we would expect two distinct patterns: 1) a monotonic increase in match rates against elicitation frequency, and 2) eventual convergence of the curves to a match rate of one.

However, not all curves exhibit a consistent monotonic increase. 
While there is a rapid initial increase in match rates across all relation types, only the synonymy curve continues to increase, ultimately converging to a match rate of one. 
The other relations stop increasing at a certain point, roughly halfway between zero and the largest elicitation frequency.
In other words, the match rate increases only at lower elicitation frequencies.

Apart from synonymy, holonymy and antonymy also reach a match rate of one.
For hypernymy, hyponymy, and meronymy, the curves even drop to a match rate of zero. 
This suggests that some very intuitive triplets with maximal elicitation frequency for these three relations are not documented in WordNet.

To sum up, the intuitiveness of triplets can hardly explain the match rate, as the two variables do not align in the expected manner.

\subsection{Mismatch Likelihood Matrix}
\begin{figure}[ht!]
    \centering
    \includegraphics[width=.8\linewidth]{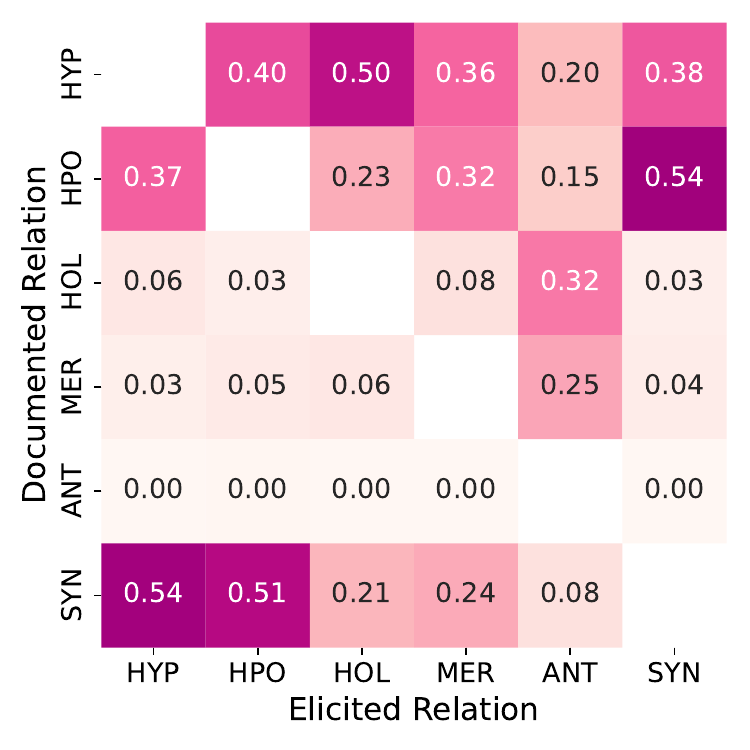}
    \caption{Mismatch likelihood matrix.}
    \label{fig:mismatched-relation}
\end{figure}

After a glance at the overall landscape, we will now look at mismatched triplets. 
Figure~\ref{fig:mismatched-relation} presents the mismatch likelihood matrix between the human-elicited relations and the WordNet-documented relations.
The vertical axis shows the documented relation $s$, and the horizontal axis shows the elicited relation $r$, where each cell gives the mismatch likelihood $\mathcal{L}(s;r)$.
Likelihoods in the same columns sum up to one.

For antonymy, the mismatch likelihood is highest for holonymy and meronymy.
Readers may recall that the mismatch rarely happens for antonymy, as we have seen in Figure~\ref{fig:catdist}.
When it happens, antonymy triplets are most likely to be documented as either holonymy or meronymy.
This mismatch often happens when a word is a metonym and involves temporal duration.
For example, (\word{day}, ANT, \word{night}) is documented in meronymy.
\word{Day} can be interpreted as \word{time for Earth to make a complete rotation on its axis} and \word{the time after sunrise and before sunset while it is light outside}, according to WordNet.
This first sense contains the second, resulting in both meronymy and antonymy between \word{day} and \word{night}.

For other elicited relations than antonymy, elicited triplets are likely to be documented as hypernymy, hyponymy, and synonymy~(high values in the first, second, and last rows).
This tendency is particularly strong when the elicited relation is either hypernymy, hyponymy, or synonymy.
It is extremely rare for triplets elicited for antonymy to be mismatched with other relations~(low values in the fifth row).

Furthermore, the semantic characteristic of words may influence the mismatch likelihood as well, and we may need an augmentation method that is sensitive to such characteristics.
For example, mismatch likelihoods could be higher for words that refer to an abstract concept rather than a physical entity~(abstract or physical word, in short).
Abstract words are often more context-dependent, making them more polysemous than physical words~\footnote{
We define abstract words as those whose all synsets are a descendent of \word{abstraction.n.06}.
Physical words are those whose all synsets are a descendent of \word{physical\_entity.n.01}.
In our data, abstract target words have more synsets~(3.2 on average) than physical target words~(2.3 on average).
The difference is statistically significant by a Mann-Whitney U test with a significance level of 5\%.}.
Because of the relatively strong polysemous nature of abstract words, we expect them to result in higher mismatch likelihoods.

We find 6,723 triplets, whose target word and relatum are abstract words and 6,555 physical triplets. 
Mismatch likelihoods are calculated for each.
For mismatch likelihood of hypernymy given hyponymy, the abstract triplets show a higher value than the physical triplets~(0.43 vs. 0.28).
For likelihoods of hyponymy given hypernymy, the abstract triplets also exceed the physical triplets~(0.43 vs. 0.31), aligning with our expectations.

\subsection{Distance of Matched Triplets}
\begin{figure}[ht!]
    \centering
    \includegraphics[width=.9\linewidth]{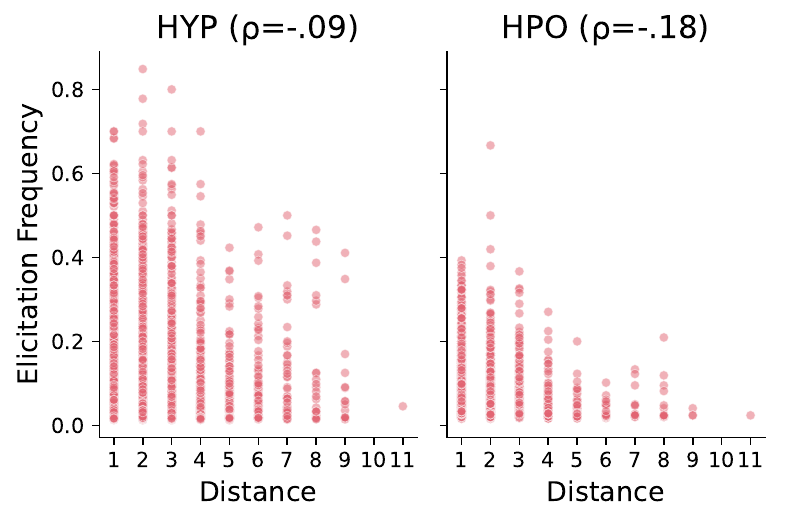}
    \caption{Distances of matched triplets.}
    \label{fig:hyp-dist}
\end{figure}
We now include indirectly matched hypernymy and hyponymy triplets in the analysis.
We define the distance of matched triplets as the length of the shortest path between the target word and relatum in the WordNet hierarchy; directly matched triplets have a distance of one.

We find that 1,895 indirectly matched triplets out of 9,890~(19\%) missing hypernymy triplets and 1,331 out of 4,338~(30\%) missing hyponymy triplets.
They are more than the directly matched 932 hypernymy triplets and 922 hyponymy triplets.

Figure~\ref{fig:hyp-dist} shows the relation between the elicitation frequency and the distance of both directly~(with a distance of one) and indirectly~(with a distance above one) matched triplets for hypernymy and hyponymy.
More than 50\% of the indirectly matched triplets have distances between two and four~(1,541 for hypernymy and 1,187 for hyponymy).
For both relations, triplets with a distance of less than four have similar ranges of elicitation frequency.

We calculate the Spearman'$\rho$ between the distance and elicitation frequency of the triplets.
The $\rho$ values are $-0.09$ for hypernymy and $-0.18$ for hyponymy,
indicating a negligible correlation between WordNet path length and language users' intuitiveness.
We conclude that the WordNet path length is not an indicator of language users' intuition on hypernymy and hyponymy triplets.
Previous literature reports a related phenomenon: humans tend to find indirect hypernymy or hyponymy triplets more intuitive compared to direct triplets~\citep{hyperlex}.

The low correlations between the elicitation frequency and the WordNet path length could be attributed to the difference in their information carried.
The direct relatum in WordNet is identified manually by lexicographers, which usually lexicalizes the generic term included in the definition~\citep{Fellbaum_1998}.
In this regard, the path distance carries the information of the difference in the degree of generality between a hypernym and a hyponym.
However, the elicitation frequency carries the intuitiveness to what extent the relatum is related to the given target word in the given relation.
Intuitiveness involves a handful of factors beyond the difference in generality so that the elicitation frequency and the path length are not in a linear relation.

For example, intuitive hypernyms or hyponyms could be the words that lexicalize a basic level category~\citep{Rosch_1976}.
They are considered to hold a centric position in the cognitive abstraction of humans and hence are easy to recall.
However, the basic level categories include two factors: they not only best generalize the similarity shared among their instances but also are most differentiated from the other categories.
In this regard, they do not just minimize the difference in generality as the direct hypernym or hyponym.

\section{Conclusion}
In the present work, we provide a straightforward and flexible methodology of comparison between language users' intuition and WordNet as the preliminary step of systematic augmentation of WordNet.
Our findings suggest that a misalignment exists between them; it can be observed from the following aspects.
First, the majority of elicited triplets are overall missing in WordNet, regardless of relations; even highly intuitive triplets could be missing in WordNet. 
Second, for some word pairs, there is a mismatch between the elicited relation and the WordNet-documented relation.
Finally, WordNet path length is not an indicator of language users' intuition.
This misalignment suggests the needs and directions of augmentation.

\section{Future Work}
WordNet has rich and fine-grained lexico-semantic information, which may facilitate mining missing relations.
For example, previous work~\citep{Boyd-Graber_2005, Maziarz_2020} uses WordNet glosses to establish evocation relations, which are semantic associations where a lexical item brings another to mind.
We hypothesize the glosses might also be useful in recognizing the semantic relations that we discussed in the present work.
As a preliminary experiment, we explore whether the similarity between glosses can differentiate related~(matched and missing) and unrelated triplets.

We calculate the gloss-based similarity of word pairs in matched, missing, and unrelated triplets and compare the similarities among triplet groups.
The gloss-based similarity for triplets is defined as the maximum BERTScore~\citep{bertscore} computed across all possible gloss pairs between the target word and relatum.
The gloss-based similarity ranges between zero and one, with higher values indicating a greater degree of similarity between the glosses of two words.

We use non-hapax triplets for matched and missing triplets~(15,806 in total).
30,000 unrelated triplets are sampled from WordNet, guaranteeing both the target word and relatum in the triplets appear in the human elicitation data.
A Mann-Whitney U test at a significance level of 5\% is then performed between the unrelated triplets against matched or missing triplets within every relation.
We additionally apply the test between matched and missing triplets.
We expect the gloss-based similarity for matched and missing triplets to be higher than that of unrelated triplets.

\begin{figure}[htbp]
    \centering
    \includegraphics[width=0.9\linewidth]{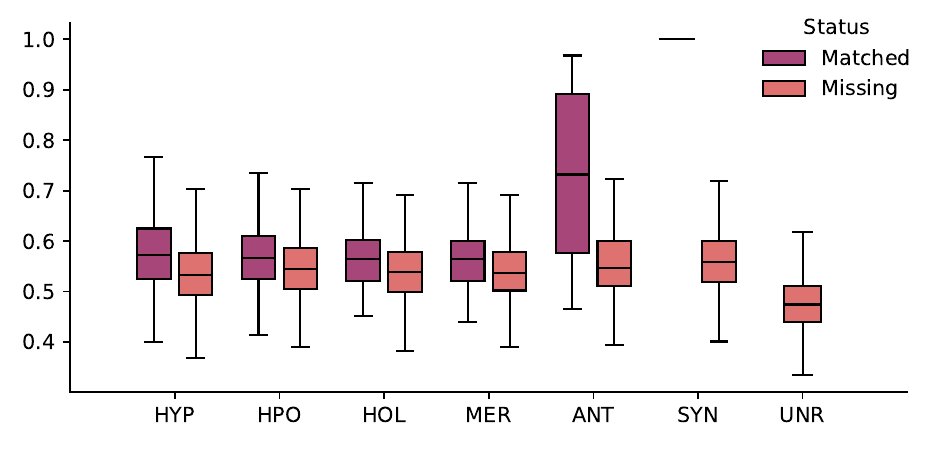}
    \caption{Gloss-based similarity of triplets per relation. UNR means unrelated triplets.}
    \label{fig:bertscore}
\end{figure}
Figure~\ref{fig:bertscore} shows the results.
For all relations, both matched and missing triplets yield significantly higher gloss-based similarities than the unrelated triplets (UNR), aligning with our expectations.
For matched triplets, the highest mean gloss-based similarities are observed for synonymy~(1.00) and antonymy~(0.74).
As two synonyms share the same gloss by definition, their similarity always marks the full value.
Two antonyms differ in only a few specific semantic features while sharing other features, leading to similar glosses.

However, the missing triplets, which are the focus of augmentation, show significantly lower similarities than the matched triplets across all relations. 
It indicates the difficulty in mining missing triplets at the same level of confidence as for the matched triplets. 
Moreover, the mean gloss-based similarities for all relations range narrowly between 0.54 and 0.57.
It suggests that distinguishing these relations based solely on the gloss-based similarity is difficult.

In summary, glosses can be useful in recognizing semantic relatedness and hence it is possible to augment WordNet.
However, relying solely on glosses is insufficient to determine whether two words have a relation, nor to identify the relation type.
To achieve an effective augmentation, it is essential to employ other information in WordNet.

\section{Limitation}
In the present study, we employed only four participants for each template. 
Their backgrounds and experiences could influence their responses, resulting in a potential bias in the experimental results.
More participants might be necessary to mitigate annotator bias. 

Another limitation is on parts of speech.
In this study, we aimed to include as many relations as possible, which narrows the coverage of parts of speech to only nouns.
However, antonymy in WordNet is mainly defined between adjectives, and hence, our results cannot be applied to the adjective antonymy triplets.

We also need to mention a methodological limitation.
In elicitation experiments, we presented the participants with the target words without any word sense disambiguation.
This led to difficulty in interpreting the results as we could not identify the word senses of the target word and the relatum in an elicited triplet.
The experiments need to incorporate a device for word sense disambiguation.
For example, one could provide a context that specifies the sense of the target word before asking relata.
In order to figure out in which sense the relata are elicited, one could also ask the participants to produce an example sentence of each relatum instead of a single word.

\section*{Acknowledgments}
This work was supported by JST SPRING, Grant Number JPMJSP2106. The third author was supported by Institute of Science Tokyo~(formerly Tokyo Institute of Technology)'s World Research Hub Program. 

\bibliography{custom}

\clearpage
\appendix
\onecolumn
\section{All Templates}
\label{app:templates}
\begin{table*}[hb!]
    \centering
    \begin{tabular}{l|l}
    \toprule
    Relation & Template \\
    \midrule
        \multirow{7}{*}{HYP~(7)}
        & a {\W} is a type of a {\V} \\
        & a {\W} is a kind of a {\V} \\
        & the word {\W} has a more specific meaning than the word {\V} \\
        & a {\W} is a {\V} \\
        & a {\W} is a specific case of a {\V} \\
        & a {\W} is a subordinate type of a {\V} \\
        & the word {\W} has a more specific sense than the word {\V} \\
        \midrule
        \multirow{4}{*}{HPO~(4)} 
        & my favorite {\W} is a {\V} \\
        & a {\W}, such as a {\V} \\
        & the word {\W} has a more general meaning than the word {\V} \\
        & the word {\W} has a more general sense than the word {\V} \\
        \midrule
        \multirow{7}{*}{HOL~(7)}
        & a {\W} is a component of a {\V} \\
        & a {\W} is a part of a {\V} \\
        & a {\W} is contained in a {\V} \\
        & a {\W} belongs to constituents of a {\V} \\
        & a {\W} belongs to parts of a {\V} \\
        & a {\W} belongs to components of a {\V} \\
        & a {\W} is a constituent of a {\V} \\
        \midrule
        \multirow{6}{*}{MER~(6)} 
        & constituents of a {\W} include a {\V} \\
        & components of a {\W} include a {\V} \\
        & parts of a {\W} include a {\V} \\
        & a {\W} consists of a {\V} \\
        & a {\W} has a {\V} \\
        & a {\W} contains a {\V} \\
        \midrule
        \multirow{9}{*}{ANT~(9)}
        & it is not likely to be both a {\W} and a {\V} \\
        & a {\W} is the opposite of a {\V} \\
        & the word {\W} has an opposite sense of the word {\V} \\
        & it is impossible to be both a {\W} and a {\V} \\
        & the word {\W} has a meaning that negates the meaning of the word {\V} \\
        & it is a {\W} so it is not a {\V} \\
        & the word {\W} has an opposite meaning of the word {\V} \\
        & if something is a {\W}, then it can not also be a {\V} \\
        & the word {\W} has a sense that negates the sense of the word {\V} \\
        \midrule
        \multirow{7}{*}{SYN~(7)}
        & a {\W} is also known as a {\V} \\
        & a {\W} is often referred to as a {\V} \\
        & the word {\W} has a similar meaning as the word {\V} \\
        & a {\W} is similar to a {\V} \\
        & the word {\W} means nearly the same as the word {\V} \\
        & a {\W} is indistinguishable from a {\V} \\
        & a {\W} is also called a {\V} \\
    \bottomrule
    \end{tabular}
    \caption{All templates used in data collection are presented by relation.}
    \label{tab:all_templates}
\end{table*}

\clearpage
\section{Distributions of hapax and non-hapax triplets.}
\label{app:distinguished-dist}
\begin{figure*}[hp!]
    \centering
    \includegraphics[width=.8\linewidth]{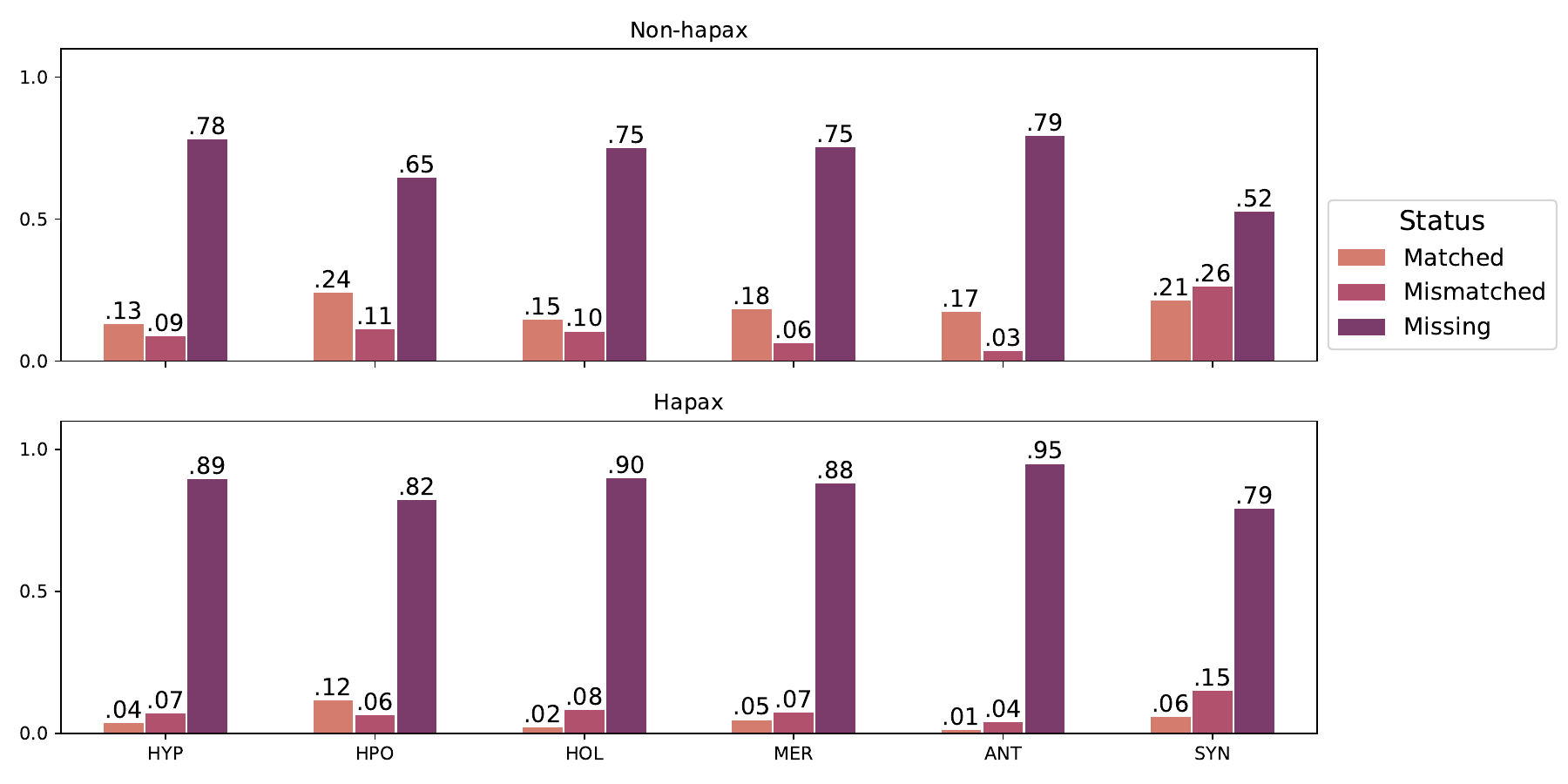}
    \caption{Match status distribution per relation, distinguishing hapax and non-hapax triplets.}
    \label{fig:distinguished-dist}
\end{figure*}

\end{document}